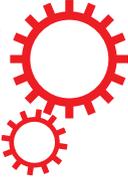

# SCIENTIFIC REPORTS

OPEN

# Algorithm guided outlining of 105 pancreatic cancer liver metastases in Ultrasound

Alexander Hann[1,2], Lucas Bettac[1], Mark M. Haenle[1], Tilmann Graeter[3], Andreas W. Berger[1], Jens Dreyhaupt[4], Dieter Schmalstieg[5], Wolfram G. Zoller[2] & Jan Egger[5,6]

Manual segmentation of hepatic metastases in ultrasound images acquired from patients suffering from pancreatic cancer is common practice. Semiautomatic measurements promising assistance in this process are often assessed using a small number of lesions performed by examiners who already know the algorithm. In this work, we present the application of an algorithm for the segmentation of liver metastases due to pancreatic cancer using a set of 105 different images of metastases. The algorithm and the two examiners had never assessed the images before. The examiners first performed a manual segmentation and, after five weeks, a semiautomatic segmentation using the algorithm. They were satisfied in up to 90% of the cases with the semiautomatic segmentation results. Using the algorithm was significantly faster and resulted in a median Dice similarity score of over 80%. Estimation of the inter-operator variability by using the intra class correlation coefficient was good with 0.8. In conclusion, the algorithm facilitates fast and accurate segmentation of liver metastases, comparable to the current gold standard of manual segmentation.

Pancreatic ductal adenocarcinoma (PDAC) has a poor prognosis, despite new developments regarding chemotherapy regimen in the last years, with a survival rate of less than 7% of all patients five years after diagnosis[1]. Additionally this disease is projected to become the second most common cause of cancer death in 2030[2]. Most of the patients are diagnosed with a metastatic disease stage with no option for a surgical resection in curative intent[3]. New therapeutic options that were introduced in the recent years prolong survival of these patients diagnosed in a metastatic state[4,5]. The establishment of second line therapies even increase the duration of treatment[6–9]. Similarly, the second most common carcinoma arising from the pancreas, the neuroendocrine neoplasms (NEN), are also associated with a poor prognosis when diagnosed with liver metastasis[10].

Due to the prolonged time of these patients receiving chemotherapy, the number of staging examinations, which are done every two to three months, is also growing[11]. Ultrasound (US) is a cheap and fast method to visualize focal liver lesions. It has a similar sensitivity regarding focal liver lesions in direct comparison to computed tomography (CT) or magnetic resonance tomography (MRI). Ultrasound is - probably due to the mentioned benefits - more widespread than CT or MRI and thus included as the imaging modality of choice in the German guideline for follow up of patients after resection of stage I or II colon cancer[12]. The ESMO – ESDO clinical practice guidelines recommend US for the response evaluation of pancreatic cancer patients undergoing palliative treatment[13]. Due to the fact that liver metastasis have different appearances in ultrasound, the interpretation of these images has a poor inter-observer agreement[14,15]. The outlining and segmentation of liver lesions for evaluation of their size often results in different values due to this poor inter-observer agreement. Semiautomatic segmentation algorithms promise a faster and more accurate way to segment lesions but are often performed on a small set of lesions by examiners that often worked with the algorithm before. Still computer aided automatic delineation of masses are successfully used for different image modalities like in radiation therapy using MRI images[16] or in the detection of breast nodules in breast cancer screening using ultrasound[17,18].

[1]Department of Internal Medicine I, Ulm University, Ulm, Germany. [2]Department of Internal Medicine and Gastroenterology, Katharinenhospital, Kriegsbergstraße 60, 70174, Stuttgart, Germany. [3]Department of Diagnostic and Interventional Radiology, Ulm University, Ulm, Germany. [4]Institute of Epidemiology & Medical Biometry, Ulm University, Ulm, Germany. [5]Institute for Computer Graphics and Vision, Graz University of Technology, Inffeldgasse 16, 8010, Graz, Austria. [6]BioTechMed, Krenngasse 37/1, 8010, Graz, Austria. Correspondence and requests for materials should be addressed to A.H. (email: alexander.hann@uniklinik-ulm.de)

Using the medical platform MevisLab[19], we implemented a computer-based method to measure focal liver lesions on B-mode US images via a semiautomatic method. The fundamental method has already been successfully applied to the segmentation of pituitary adenomas[20], prostate central glands (PCG)[21], vertebral bodies[22], aneurysms and stenosis[22], and even liver ablation zones[23]. A detailed description of translating the basic method for the application of US images has been previously reported by the authors[24]. In this study however, we aim to evaluate the method by users that never worked with the algorithm before and on a much larger set of data. We retrospectively identified 105 B-mode US images of liver metastasis generated during the treatment course of 56 patients with histologically confirmed pancreatic cancer. Two examiners outlined in a first session all lesions and determined the greatest diameter. In a second session after five weeks, the algorithm was used to measure the lesions in a random order. The examiners were given up to five minutes to measure ten images in order to learn how to use the program. Afterwards, they were asked to measure the 105 metastases using the algorithm. In summary, the two examiners were satisfied with the result of the semiautomatic segmentation in 92 and 94 of the 105 cases. Regarding these measurements, the segmentation of the liver metastasis took median 17.2 and 10.2 seconds using the manual method, compared to 9.5 and 8.2 seconds using the semiautomatic method. The median differences of manual and semiautomatic segmentations were significant ($p < 0.01$) in both cases. Additionally, we were able to measure the overlap of the areas measured with the manual and semiautomatic method using the Dice similarity scores (DSC)[25] and the Hausdorff Distances (HD)[26]. This resulted in a median DSC of 84% (95% CI; 82.4–85.3) for examiner 1 and 82% (95% CI; 80.3–84.3) for examiner 2. The median HD were 9 (95% CI; 8.6–10.6) and 10 (95% CI; 9.4–11.2) pixels, respectively.

## Materials and Methods

**Data Acquisition.** Ultrasound examinations were performed using a multifrequency curved probe, which allows ultrasound acquisitions with a bandwith of 1 to 6 MHz (LOGIQ E9/GE Healthcare, Milwaukee, Il, USA and Toshiba Aplio 80, Otawara, Japan). Using the digital picture archive of the ultrasound unit at the Katharinenhospital Stuttgart (Germany), we retrospectively selected images of liver metastasis obtained from patients with PDAC or NEN. The images were included, if they fulfilled the following criteria: visible metastases surrounded by liver tissue with no visible marker or text overlaying the target lesion. Patient information was removed from the image, and the anonymized picture was subsequently ordered randomly. The local ethics committee (Ethik-Kommission der Landesärztekammer Baden-Württemberg, Stuttgart, Germany) provided a waiver of the requirement for informed consent for this retrospective study and allowed the publication of anonymized data.

**Segmentation Method.** The segmentation method applied in this evaluation study belongs to the class of graph-based algorithms[27] and the section follows the explanations in ref.[24]. Graph-based algorithms convert an image or parts of it into a graph $G(V, E)$ that consists of nodes $n \in V$ and edges $e \in E$. The nodes $n \in V$ are sampled in the image joined by two virtual nodes $s \in V$ and $t \in V$, which are called the source and the sink, respectively. The two virtual nodes $s$ and $t$ are utilized to perform an s-t-cut[28] after the graph construction. In doing so, the graph is split in two parts: the foreground (which could be the liver metastasis) and the background (which could represent the remaining surrounding liver tissue). Moreover, the edges $e \in E$ establish direct connections between the sampled and two virtual nodes, like an edge $<v_i, v_j> \in E$ that connects the nodes $v_i, v_j$[29]. In our approach the graphs' nodes are sampled along radial rays, which are equidistantly distributed around a fixed point in a clockwise manner (Supplementary Figure 1). After the nodes have been sampled, the so called $\infty$-weighted intra-edges that connect nodes along the same ray are set up (Supplementary Figure 2). In summary, the $\infty$-weighted intra-edges ensure that the s-t-cut affects only one of the edges that belong to the same ray, which ensures a star shaped min-cut/max-flow result[30]. Following, the $\infty$-weighted inter-edges between nodes from different rays are set up (Supplementary Figures 3 and 4). Here, a delta value $\Delta_r$ influences the number of possible s-t-cuts, hence influencing the smoothness of the segmentation result. In practice, the greater the delta value, the greater is also the flexibility of the segmented contour. However, for our evaluation study, we set the delta value to a fixed value of two ($\Delta_r = 2$). In the ensuing step, the weighted edges between the nodes and the source/sink are set up. Therefore, an average gray value is needed and sampled on the fly around the user-defined seed point. The average gray values is used to determine the absolute differences between this average gray value and the gray values behind the sampled nodes. Last but not least, the differences between two adjacent absolute values are calculated, which finally represent the weights of the edges. Note, that the signs of the differences (negative/positive) define if a node is bound to the source or the sink, with the exception for the very first and last nodes of every ray, which are bound with their absolute weight values to the source or sink, respectively. In fine, our graph construction replaces the pre-defined templates with fixed seed point positions used in previous works[21,31,32], with a circular template centered around a user-defined seed point in the US image.

**Measurement.** Manual and semiautomatic measurement was performed on a Lenovo Yoga 2 Pro laptop with an Intel Core i7-4500U CPU @ 2.40 GHz, 8 GB RAM and Windows 8.1, 64 bit installed. Two examiners with over 10 years-experience in performing and interpreting over 20,000 ultrasound examinations performed the measurements. The chosen images were never tested with the algorithm nor were the images known to the examiners. Additionally, the examiners had never used the algorithm before.

In the first session, both examiners outlined each of the metastasis manually and measured the largest diameter (diameter a), with an additional shorter diameter 90 degree related to the first one (diameter b). Time of measurement per metastasis was recorded. In the second session, after five weeks to decrease bias due to memory effect, the examiners performed the semiautomatic measurement. The order of the metastasis during this session was randomly redistributed. A short training including additional 10 images of pancreatic cancer liver metastases images was performed for up to five minutes prior to the session to exercise the usage of the algorithm.

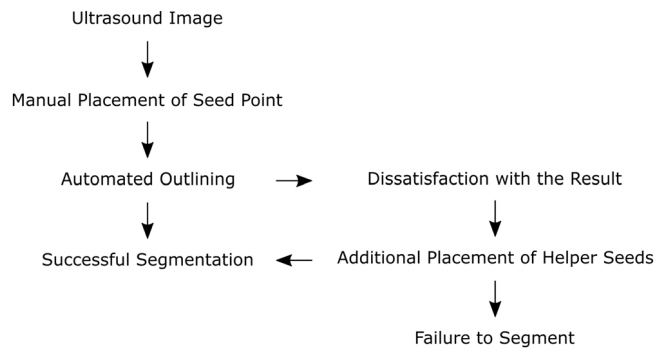

**Figure 1.** Overview of the semiautomatic segmentation of the liver metastasis using US-CUT.

| Median (mm) | Q$_1$ (mm) | Q$_3$ (mm) | Min (mm) | Max (mm) |
|---|---|---|---|---|
| 20 | 14 | 27 | 6 | 115 |

**Table 1.** Length of the largest diameter of the 105 metastases. Q$_1$ and Q$_3$ = quartile 1 and 3.

The algorithm was applied as described before[24]. First, a seed point was applied in the middle of the liver lesion, which could additionally be improved by dragging the seed point around (inside the lesion). The algorithm immediately presents segmentation contours around the edges of the lesion. If the examiner is satisfied, the next image was loaded. In case of dissatisfaction with the outlining, additional helper seeds were placed by the examiner to alter the morphology of the outlining (Fig. 1). Note: the placement of one or more additional helper seeds still allowed the dragging of the initial seed point in the middle of the liver lesion, only with the difference that the helper seeds remain on the lesions border, thus supporting the algorithm with this additional knowledge (location/gray values). Dissatisfaction with the result of the outlining despite additional helper seeds and time spent on segmentation were noted. We measured the time spent per metastasis and calculated the degree of area overlap, comparing manual and semiautomatic segmentations using the Dice similarity score (DSC)[25]. A value of 100% resembles perfect overlap of two areas. Hausdorff distance (HD) was used in addition to the DSC to measure the degree of overlap in pixels[26]. This distance resembles the greatest of all distances from a point in one segmentation to the closest point in the other segmentation. Additionally, we calculated the difference between the two manually measured greatest diameters (a and b) and automatically calculated diameters from the algorithm after semiautomatic segmentation in millimeter. Statistics were performed with SPSS 24. The 95% confidence interval of the median was calculated using Bootstraping. The Wilcoxon rank sum test was used to investigate differences between time spent for manual and semiautomatic segmentation. Additionally, the two-sided one sample student t-test was used to test a significant difference between manual and semiautomatic time per examiner. The intra class correlation coefficient (ICC) of the DSC (manual vs. semiautomatic) of examiner 1 and 2 was used to estimate the inter-operator variability.

## Results

### Characteristics of the selected images.
Selection of retrospective images obtained during the treatment of patients with pancreatic cancer yielded 77 ultrasound images of 46 patients diagnosed with PDAC. Additionally, 28 images of 10 patients with neuroendocrine pancreatic neoplasia were identified. All images displayed liver metastases without overlaying of marker or text. The median of the maximal lesion diameter was 20 mm (Table 1).

The examiners were satisfied in 92 (88%) and 94 (90%) of the 105 cases with the result using the algorithm. In seven identical cases, both examiners were not satisfied with the outlining drawn by the algorithm despite the placement of helper seeds. In the following sections, our analysis will focus on the measurements of the segmentation that were marked as "satisfied", unless we explicitly mention comparisons to the segmentations that were marked "not satisfied."

### Comparison of time spent for manual and semiautomatic segmentation.
Analyzing the amount of time each examiner needed for the manual segmentation per image revealed that examiner 1 (n = 92) spent median 17.2 seconds, compared to 10.2 seconds spent by examiner 2 (n = 94). Regarding the manual segmentation, examiner 2 was faster, but both examiners needed less time, when they were using the semiautomatic algorithm. (Fig. 2) The semiautomatic segmentation was in median 8 seconds faster when performed by examiner 1 and 2.2 seconds faster when performed by examiner 2. Both differences were statistically significant using the Wilcoxon rank sum test regarding (p < 0.01) (Table 2). The mean values were 8.8 seconds and 2.0 seconds for examiner 1 and 2 respectively. These values were as well significant using the two-sided one sample student t-test (p < 0.01). Interestingly, analyzing the dataset including the segmentations marked as "not satisfied" revealed that examiner 1 had images where he spent more time on the semiautomatic segmentation than on the manual outlining. All these cases were part of the segmentations marked as "not satisfied". This reflects the additional amount of time spent on trying to achieve a good result using the algorithm.

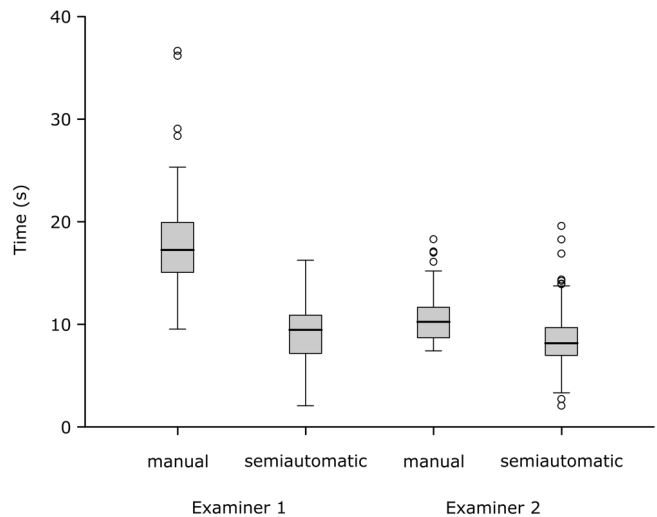

**Figure 2.** Comparison of time spent on manual and semiautomatic measurement. Box-and-whisker plots illustrate 92 and 94 marked as satisfied measurements by examiner 1 and 2 respectively.

|  | Manual (s) | | | | | Semiautomatic (s) | | | | | |
| --- | --- | --- | --- | --- | --- | --- | --- | --- | --- | --- | --- |
|  | Median | $Q_1$ | $Q_3$ | Min | Max | Median | $Q_1$ | $Q_3$ | Min | Max | p* |
| Examiner 1 (n = 92) | 17.2 | 15.1 | 19.9 | 9.5 | 36.7 | 9.5 | 7.3 | 10.9 | 2.1 | 16.3 | p < 0.01 |
| Examiner 2 (n = 94) | 10.2 | 8.7 | 11.6 | 7.4 | 18.3 | 8.2 | 7.0 | 9.7 | 2.1 | 19.6 | p < 0.01 |

**Table 2.** Comparison time spent on measuring metastasis per image. Only semiautomatic measurements regarded as satisfied included. $Q_1$ and $Q_3$ = quartile 1 and 3. *The p-value is related to the median difference between manual and semiautomatic segmentation time.

**Comparison of manual and semiautomatic segmentation.** The DSC was used to determine the degree of resemblance between two segmentation areas that are superimposed on one another (Fig. 3). We calculated the DSC for manual compared to semiautomatic segmentation for each examiner (Table 3). It revealed a median DSC of 84% (95% CI; 82.4–85.3) for examiner 1 and 82% (95% CI; 80.3–84.3) for examiner 2. Additional, the calculated median HD revealed a low median distance of 9 pixels (95% CI; 8.6–10.6) for examiner 1 and 10 pixel (95% CI; 9.4–11.2) for examiner 2. Furthermore, in order to analyze the accuracy of the semiautomatic segmentation, the examiners were asked to manually measure the largest diameter (diameter a) with an additional shorter diameter 90 degree related to the first one (diameter b). The corresponding two diameters were calculated by the algorithm automatically during the semiautomatic outlining using the resulting contour of the lesion. The difference between manual and semiautomatic segmentation revealed a median of only 2 mm ($Q_1$-$Q_3$; 1–4) for diameter a and 1 mm ($Q_1$-$Q_3$; 0–2) for diameter b measured by examiner 1. Examiner 2 had a median difference of 3 mm ($Q_1$-$Q_3$; 2–5) for diameter a and 2 mm ($Q_1$-$Q_3$; 1–3) for diameter b (Supplementary Table 1). Estimation of the inter-operator variability using the ICC of the DSC manual vs semiautomatic in segmentations marked a satisfied by both examiners (n = 88) was 0.8.

**Segmentations regarded as not adequate by the examiners.** Examiner 1 regarded 13 segmentations and examiner 2, 11 segmentations as not good enough. Seven of these were regarded by both as inadequate. Supplementary Figure 5 presents an example of a segmentation regarded by examiner 1 as inadequate compared to the segmentation of the same image by examiner 2, who was satisfied with the result. Examiner 1 defined the echo-poor halo surrounding the metastasis as part of the lesion. Unfortunately, the outlining of the semiautomatic segmentation does not reach the manually drawn line despite the use of helper seeds. In contrast, examiner 1 regarded a thinner halo as part of the lesion and was satisfied with the semiautomatic segmentation result. A possible explanation for the result of examiner 1 is that the homogenous thick halo of this metastasis is regarded as normal liver tissue by the algorithm.

## Discussion

In this contribution, we presented the results of a fast, interactive segmentation algorithm for pancreatic liver metastases in ultrasound images. We decided to evaluate this new method using a large number of images displaying hepatic metastases of pancreatic cancer patients, which have never before been tested with this algorithm. Additionally, we asked two examiners to participate in the evaluation, who had never used the algorithm, and gave them only up to five minutes to learn how to use it. Despite these challenges, the algorithm performed very well with a DSC of over 80% and a 95% CI starting at over 80%. The differences in the diameters of 1 to 3 mm were very small, and both examiners were satisfied with the results in nearly 90% of the cases. Finally, the semiautomatic measurement was significantly faster than the manual one in both cases. Although the median difference in the manual and semiautomatic segmentation time was just 8 and 2.2. seconds, these differences represent 47% and 22% of the manual segmentation time respectively. The time during segmentation represents a critical step in examination due to the high mental concentration of the examiner to carefully segment the lesion without over or underestimation of its size.

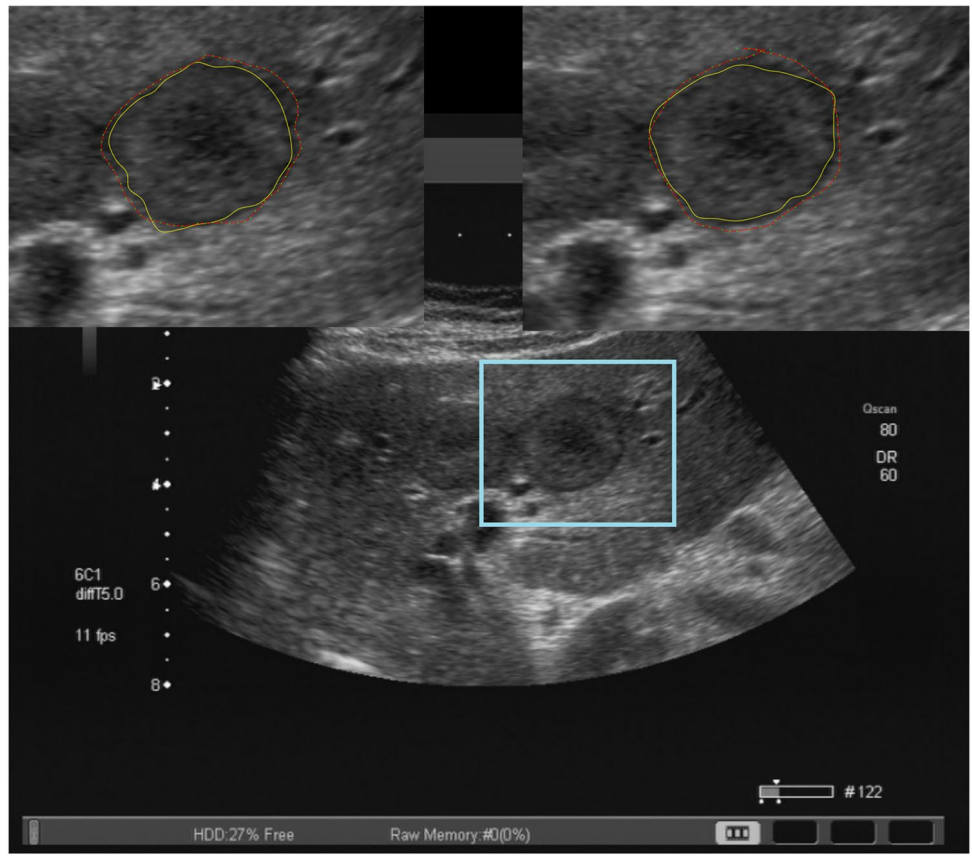

**Figure 3.** Example of a segmentation. Depicted are the native image in the background and two zoomed-in views of the metastasis (representing the blue box). The upper left box represents the segmentation results of examiner 1 and the upper right box the results of examiner 2. The red outlines represent the manual segmentations and the yellow outlines represent the results of the semi-automatic segmentation.

|  | Area DSC (%) | | | | Area HD (Voxel) | | | |
| --- | --- | --- | --- | --- | --- | --- | --- | --- |
|  | Median | 95% CI | Min | Max | Median | 95% CI | Min | Max |
| Examiner 1 (n = 92) | 84 | 82.4–85.3 | 63 | 93 | 9 | 8.6–10.6 | 4 | 31 |
| Examiner 2 (n = 94) | 82 | 80.3–84.3 | 57 | 94 | 10 | 9.4–11.2 | 4 | 133 |

**Table 3.** Comparison of satisfied manual and semiautomatic measurement per examiner. DSC = Dice similarity score, HD = Hausdorff distance, CI = confidence interval.

A search of the literature revealed different publications in peer reviewed journals and contributions to congresses that described methods for segmentation of ultrasound images. Most of these publications discuss the difficulties of segmentation performed using ultrasound images and propose new ways to overcome difficulties like speckle noise existence[33]. One way is to enhance the contrast of the lesion in comparison to the surrounding tissue by using additional intravenous contrast dye[34–36]. This time-consuming method truly helps to delineate focal liver lesion, but harbors the risk of allergic reaction to the contrast dye and significantly increases the costs and length of the examination. The latter disadvantage is due to the three contrast phases that are usually intermittent recorded during an examination and that usually take up to 6 minutes[37]. In regular follow-up visits of cancer patients, only the sizes of liver metastases are assessed using native non-contrast enhanced images.

One study involved 55 cases of focal liver lesions that received a contrast enhanced ultrasound (CEUS) for workup of a suspected liver lesion[35]. After selection of a key frame derived from the CEUS examination video, the radiologist placed a seed point that induced the segmentation process. The reference in this study was a manual segmentation of the liver lesion done by one radiologist. Although using the proposed algorithm resulted in a good segmentation result, only 11 out of 55 images were segmented by two radiologists. Additionally, it is not clear if the chosen images were directly evaluated for the first time with the proposed algorithm. Studies that proposed native, non-contrast enhanced ultrasound images often involved just a few cases to present the feasibility of the method[38,39]. One study that evaluated a semi-supervised segmentation algorithm using different ultrasound datasets of focal liver lesions achieved very good Dice scores of over 90% compared to an expert radiologist who manually segmented the focal liver lesions[40]. This study involved images obtained from 15 patients. Although faster than the previously proposed algorithm by the group, the time for semi-automated segmentation ranged from 18 to 35 seconds, with a median of 27 seconds. This is slower than the time spent using our algorithm that was 10 seconds for examiner 1 and 8 seconds for examiner 2.

Finally, there is some previous work from the authors that is related to this contribution, In our previous work[23,41,42], we apply and evaluate a semi-automatic segmentation algorithm for radiofrequency ablation (RFA) zones via optimal s-t-cuts. The interactive graph-based approach builds upon a polyhedron to construct the graph. However, it was specifically designed for computed tomography acquisitions from patients that had RFA treatments of hepatocellular carcinomas (HCC). For an evaluation, we used twelve post-interventional CT datasets from the clinical routine and, as evaluation metric, we utilized also the Dice score. Compared with pure manual slice-by-slice expert segmentations from interventional radiologists, we were able to achieve a Dice score of about 80%, and our approach was even able to handle images containing the RFA needles still in place. RFA of liver tumors induces areas of tissue necrosis, which can be visualized reliably in contrast enhanced CT. This results in a bright border around the ablation zone, which makes an (automatic) segmentation easier and more reliable. The same applies for brain tumors, like glioblastoma multiforme in contrast-enhanced MRI acquisitions, where a bright border is surrounding a core of necrotic brain tissue[32,43,44].

Although this study presents encouraging results for the application of the algorithm in clinical routine, there are still some limitations that need to be addressed. We evaluated the algorithm with two examiners. Evaluating the method with more physicians would have increased statistical plausibility. Due to the missing direct access to video-output of an ultrasound machine, the study had to be performed with retrospectively collected images. This was also necessary because of the aim to include more than 100 images of different metastasis. We included only pancreatic cancer patients, because, in our clinic, those are often staged during palliative treatment using abdominal ultrasound combined with endoscopic ultrasound.

There are several areas for future work, like the implementation of this tool to function on computers that receive the images directly from the ultrasound machine. It could be evaluated prospectively with images drawn from patients with different cancer entities. Additionally, we hope to implement the recognition of metastasis using a 3D ultrasound dataset. This segmentation is still very time consuming and often needs manual outlining of the same metastasis in more than four images in order to achieve a good result. The acquisition of 3D content with segmented liver metastasis might help implement such 3D ultrasound images in virtual reality environments[45] that are explored using head mounted displays[46]. These displays allow immersion into computer-created real-time simulations. They have great potential to influence the way we plan interventions and teach about different disease entities.

In summary, this work presents the use of an algorithm for segmentation of liver metastasis due to pancreatic cancer under real life conditions. We believe that implementation of such a method into daily practice will help assist the examiner without compromising accuracy.

### Acknowledgements

The work received funding from BioTechMed-Graz ("Hardware accelerated intelligent medical imaging") and the 6th Call of the Initial Funding Program from the Research & Technology House (F&T-Haus) at the Graz University of Technology (PI: Dr. Dr. Dr. habil. Jan Egger). Alexander Hann is supported by a Baustein Grant of the University Ulm.

### Author Contributions

Conceived and designed the experiments: A.H., J.E. Performed the experiments: A.H., L.B., M.M.H., T.G., A.B. Analyzed the data: A.H., J.E., J.D. Contributed reagents/materials/analysis tools: J.E., A.H., A.B., W.G.Z., D.S. Wrote the paper: A.H., J.E.

### Additional Information

**Supplementary information** accompanies this paper at https://doi.org/10.1038/s41598-017-12925-z.

**Competing Interests:** The authors declare that they have no competing interests.

**Publisher's note:** Springer Nature remains neutral with regard to jurisdictional claims in published maps and institutional affiliations.